# Model Criticism of Bayesian Networks with Latent Variables

**David M. Williamson**
The Chauncey Group Intl.
664 Rosedale Road
Princeton, NJ 08540

**Russell G. Almond**
Educational Testing Service
664 Rosedale Road
Princeton, NJ 08540

**Robert J. Mislevy**
Educational Testing Service
664 Rosedale Road
Princeton, NJ 08540

## Abstract

The application of Bayesian networks (BNs) to cognitive assessment and intelligent tutoring systems poses new challenges for model construction. When cognitive task analyses suggest constructing a BN with several latent variables, empirical model criticism of the latent structure becomes both critical and complex. This paper introduces a methodology for criticizing models both globally (a BN in its entirety) and locally (observable nodes), and explores its value in identifying several kinds of misfit: node errors, edge errors, state errors, and prior probability errors in the latent structure. The results suggest the indices have potential for detecting model misfit and assisting in locating problematic components of the model.

## 1 INTRODUCTION

This study investigates statistical methods for identifying errors in Bayesian networks[1] (BN) with latent variables, as found in intelligent educational assessments.[2] The success of an intelligent assessment or tutoring system depends on the adequacy of the *student model*, representing the relationship between the unobservable cognitive variables of interest ($\theta$s) and the observable features of task performance ($x$s), with the probability model for $x$ given $\theta$ being expressed as a BN.

The student model is constructed on the basis of a *cognitive task analysis* (CTA), an investigation of the cognitive components that contribute to task performance (Mislevy, Steinberg, Breyer, Almond, & Johnson, 1999). There is no assurance that the resulting student model is an accurate representation of the true structure of cognition, or that it is the most useful model for the purpose of the assessment. *Model criticism* means evaluating the adequacy of a statistical model, enabling the analyst to discover hypotheses, variables, or relationships beyond those represented in the original model—to improve the structure of the BN in response to mismatches between modeled and observed data patterns (Mislevy, 1994; Mislevy & Gitomer, 1996).

At present, the current process of critiquing, refining, and validating a student model depends largely on examining the model from the perspective of the findings of the CTA and from theoretical considerations of cognition in the domain. The use of statistical diagnostic tools is notably lacking. Developing and using empirical tools for model criticism, therefore, is important to the continued development and implementation of BN methodologies in cognitive assessment. Statistical indices of model fit could be useful in cognitive assessment in several ways, such as (1) comparing proposed modeled structures to preliminary performance CTA data; (2) evaluating the model-data concordance for nodes upon which examinee classification decisions are based, (3) identifying examinee performance that is inconsistent with the posited student model, and (4) confirming the appropriateness of the modeled cognitive structure and, by implication, providing evidence about the validity of that conceptualization of cognition in the domain.

This paper focuses on the quality of the probability model for $x$ given $\theta$. Section 2 describes potential BN latent structure errors, and Sections 3 and 4 describe the methodology and results of our study. Section 5 provides discussion and directions for future research.

---

[1] Usually referred to as Bayesian inference networks (BIN) for applications in intelligent educational assessments

[2] The interested reader is referred to the following sources for discussions of other aspects of the research program from which this work arises: cognitive psychology (Frederiksen, Mislevy, & Bejar, 1993; Nichols, Chipman, & Brennan, 1995; Steinberg & Gitomer, 1996); computer-based simulations and constructed-response tasks (Bejar, 1991; Williamson, Bejar, & Hone, 1999); probability-based reasoning (Almond, & Mislevy, 1999; Almond et al., 1999); and assessment design (Mislevy et al., 1999; Mislevy, Steinberg, & Almond, 1999).

## 2  MODELING ERRORS

This study investigates the potential of a methodology for model fit indices for identifying errors in the latent structure of a BN. In many expert systems one can criticize a BN by using it predictively, then evaluating the accuracy of the predictions by collecting actual data on any node in the network. This is not possible for the latent nodes of a cognitive assessment BN; model criticism must in this case rely on evidence from observable nodes (as is the case in factor analysis and item response theory as well). We examined the utility of several indices for identifying specific types of latent structure errors in a particular hypothetical BN model. Classes of errors that degrade a model's utility or its fidelity to cognitive processes are node errors, directed edge errors, state errors, and prior probability errors.

The two node errors we address are node over-inclusion and node under-inclusion. Node over-inclusion occurs when a given $\theta$ does not contribute to the state of any $x$ or is redundant given the other $\theta$s in the model. Node under-inclusion occurs when an important and relevant node has been omitted. Node errors are a particular concern in cognitive assessment not only because of their importance, but because they are inherently unobservable. This contrasts with more typical applications of BNs in expert systems that include only variables which are observable, at least in principle.

Directed edge errors can also be described as over-inclusion or under-inclusion. The former is including an edge from a $\theta$ to an $x$ that is not needed, while the latter is omitting an edge that is needed. The severity of this type of error depends on the relative strength of the edge that is erroneously included or omitted.

Variable state errors are subtler than node and directed edge errors, but can also be described in terms of unwarranted expansion or reduction. The $\theta$s in the BNs we consider are categorical variables. Errors occur when the number of states defined for a given $\theta$ is either greater than or less than the optimal number; i.e., an expansion error or a reduction error.

The final type of error we address lies in the specification of the conditional probabilities for an $x$ given that the correct $\theta$s have been specified as its parents. These errors can vary from extreme misspecification to mild misspecification, depending on how much the probabilities deviate from the actual distribution of the states of $x$.

These error types (node, directed edge, state, and prior probability) are hierarchical[3] in that a node error contains corresponding directed edge, state, and prior probability errors, since the missing or extraneous node includes these elements. The conceptual severity[4] of these errors with regard to the student model as a model of cognition is similarly hierarchical.

## 3  METHODOLOGY

### 3.1  INDICES

This study examined three indices, Weaver's Surprise Index (Weaver, 1948), Good's Logarithmic Score (Good, 1952), and the Ranked Probability Score (Epstein, 1969)[5], that have been used to evaluate the accuracy of probabilistic predictions in weather forecasting (Murphy & Winkler, 1984). Each measures of the degree of "surprise" felt when a datum is observed.

### 3.1.1  Weaver's Surprise Index

Weaver (1948) developed the Surprise Index to distinguish a "rare" event from a "surprising" event. An event is *surprising* if its probability is small compared with the probabilities of other possible outcomes. A *surprising* event must be a *rare* event, but a *rare* event need not be *surprising*. His definition of surprise is

$$(S.I.)_i = \frac{E(p)}{p_i} = \frac{p_1^{\,2} + p_2^{\,2} + ... + p_n^{\,2}}{p_i}, \qquad (1)$$

where there are $n$ possible outcomes of a particular probabilistic event (in BN cognitive assessments with discrete variables, $n$ possible states of a variable), $p_1$-$p_n$ are the prior probabilities of each of the $n$ possible states, $E(p)$ is the expected value of the probability, and $p_i$ is the prior probability of the observed state. Values increasingly greater than unity indicate increasingly surprising observations.

### 3.1.2  Good's Logarithmic Score

In a discussion of fees and rational decisions, Good (1952) introduced what we shall be refer to as Good's Logarithmic Score:

$$GL = \log(bp_i) \qquad (2)$$

when the (predicted) event occurs, and

$$GL = \log b(1 - p_i) \qquad (3)$$

---

[3] The exception is that state errors do not necessarily follow an edge error.

[4] Which may not be reflected in the predictive capacity of the model.
[5] The Quadratic Brier Score (Brier, 1950), Good's Logarithmic Surprise Index (Good, 1954), Logarithmic Score (Cowell, Dawid, & Spiegelhalter, 1993) and Spearman correlation coefficient were also investigated but are not discussed due to lesser promise of utility.

when it does not. Here $p_i$ is the prior probability of the event $i$ in question before making the observation, and $b$ is a penalty term that keeps a forecaster from long term gain by simply predicting the average frequency of occurrence. This penalty term is given by

$$b = -\sum_{j=1}^{r} x_j \log x_j , \qquad (4)$$

where $r$ is the number of possible outcomes and $x_j$ is the expectation of $p_j$, that is, $x_j$ is the marginal probability associated with category $j$ before the observation. Values of Good's Logarithmic Score near zero indicate accurate prediction, and values increasing from zero indicate poor prediction.

### 3.1.3  Ranked Probability Score

Epstein (1969) developed the Ranked Probability Score to evaluate forecasting accuracy when the states of the predicted variable are categories of an ordered variable (such as four categories of temperature in degrees Fahrenheit). Its distinguishing feature is that it considers how close (categorically) the predicted probabilistic outcome is to the observed outcome. The Ranked Probability Score is given by

$$S_j = \frac{3}{2} - \frac{1}{2(K-1)} \sum_{i=1}^{K} \left[ \left( \sum_{n=1}^{i} p_n \right)^2 + \left( \sum_{n=i+1}^{K} p_n \right)^2 \right] - \frac{1}{K-1} \sum_{i=1}^{K} |i - j| p_i \qquad (5)$$

where $K$ represents the number of possible outcome states and $j$ indicates the observed outcome. The Ranked Probability Score uses a linearly increasing penalty as the predicted observation becomes more distant from the observed state, implying that node categorizations are an interval scale as they progress from one extreme to the other. The values of the Ranked Probability Score vary from 0.00 to 1.00, indicating the poorest possible prediction and best possible prediction respectively.

### 3.2  THE DATA GENERATION MODEL

As a baseline for evaluating fit indices, we generated 1000 response patterns $x$ from a hypothetical BN cognitive assessment—the 'Data Generation' BN—with known nodes, edges, and conditional probabilities. Although they are simulated, we refer to these vectors as 'observed' data since they represent the data that would be observed in practice, in contrast to the $\theta$s. To calculate probabilities in BNs we used the Ergo computer program (Beinlich & Herskovits, 1990; Noetic Systems, 1996).

The Data Generation BN is a hypothetical cognitive model of ability for a general practice MD[6], as might be used to assess MD proficiency in general and as a first level of more diagnostic feedback to examinees. Table 1 describes the model variables and their possible states, and Figure 1 shows the structure of the model.

---

[6] The context of this model is provided purely for the benefit of a concrete example and is not based on a CTA nor been reviewed by a physician.

Table 1:  Data Generation Model Variables and States

| Node | Meaning | States |
|------|---------|--------|
| $\theta_1$ | Medical Ability: overall ability as a general practice MD | poor; moderate; good; excellent |
| $\theta_2$ | Pharmaceutical Ability: ability to select/prescribe medications | inappropriate; typical; precise |
| $\theta_3$ | Physical Exam: ability make appropriate observations and conclusions from physical exam | incomplete; adequate; thorough |
| $\theta_4$ | Treatment Planning: ability to develop a treatment plan for patients requiring follow-up treatment | trial-and-error; by-the-book; custom to patient |
| $X_1$ to $X_5$ | Patients 1 through 5: response to treatment (according to treatment quality) | degrade; maintain; improve; healed |

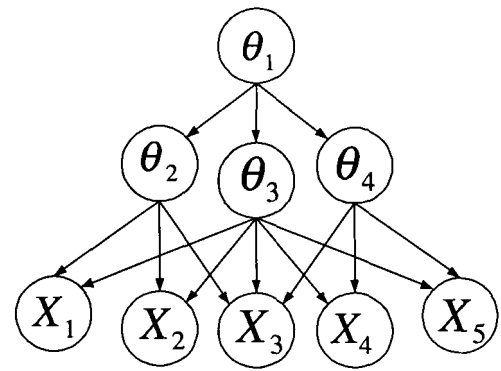

Figure 1:  Data Generation Model

The directed edges from the latent nodes to the patient-outcome nodes represent the influences of these cognitive abilities on the effectiveness of patient treatment for the cases that as they have been (hypothetically) constructed. Patients $X_1$ and $X_2$, for example, can be effectively treated on a single visit to the office with appropriate examination and medication, while patients $X_3$, $X_4$, and $X_5$ require longer-term care or repeat visits. Patients $X_4$ and $X_5$ do not require a prescription medication, but they do require repeated visits to the office. (To make this model more concrete the reader may wish to consider the simulated patients by their illness rather than as purely hypothetical cases. A skin rash or eczema might be appropriate for $X_1$, strep throat for $X_2$, a deep laceration with high chance of infection for $X_3$, partially torn ligament for $X_4$ who is a child, and influenza for $X_5$, who is an elderly patient). The $\theta_2$, $\theta_3$ and $\theta_4$ nodes have a largely conjunctive relationship (i.e. all relevant abilities must be present), meaning that an examinee must have relatively high levels of all skills a given patient requires in order to have a high probability of providing that patient with effective treatment.

## 3.3 MODEL CRITICISM COMPUTATION

A BN model can be criticized at the levels of its fit as a whole (*global measures*), of individual nodes (*node measures*), and of specified conditional probabilities (*parent-child measures*) (Spiegelhalter, et al., 1993). The latent nature of variables in cognitive assessment BNs precludes the use of parent-child measures, despite the emphasis they receive in expert systems applications (e.g. Box, 1980), and limits us to global and node measures.

Our strategy was to route predictions for observable variables through the latent structure, providing an opportunity to detect problems with the latent structure even though the student model variables could not be assessed directly. Errors in the student-model would manifest patterns of poor prediction for observable nodes individually or in the aggregate.

The 'observed data' were uploaded into the Data Generation BN. For each of the 1000 simulees, predictive probabilities were computed for each observable node treating the remaining observable nodes as known (i.e. for observable nodes $X_1$ through $X_n$ the probability that node $k$ is in state $j$ is given by $P_{kj}^* = p(X_k = j \mid X_1, \dots X_{k-1}, X_{k+1}, \dots X_n)$). The resulting probabilities for $X_2$ were treated as predictions to be compared to the observed state of $X_2$ for the simulee, as required to calculate the model criticism indices discussed above for each observed-variable node in turn for a given simulee. Carrying out this process for each of the observable nodes provided the node measures, and then aggregating across the five nodes produced a global measure for the simulee. The mean value of a node measure across the 1000 simulees served as the node measure (node-data fit) for the node in question, while the mean global measure value across the 1000 simulees served as the global measure of the model-data fit[7].

## 3.4 ERROR MODELS

The ability of model criticism indices to detect errors in the latent structure of the BN network was investigated under several conditions, each emphasizing a particular type of error: node error, directed edge error, variable state error, and prior probability error. The study was conducted in hierarchical sequence with the node error investigated first, followed by directed edge error, variable state error, and prior probability error, respectively.

### 3.4.1 Node Errors

The first stage of the study investigated the erroneous exclusion (Figure 2[8]) and inclusion (Figure 3) of a node in the student model. For each error model the conditional probabilities utilized were logical manipulations of the Data Generation model according to a possible misconception (on the part of the modelers) of the cognitive processes of medicine. The erroneous exclusion model eliminates $\theta_4$, while the erroneous inclusion model adds a node ($\theta_5$) that represents an MD's Internship Location as a contributing factor in their ability to treat patients 4 and 5.

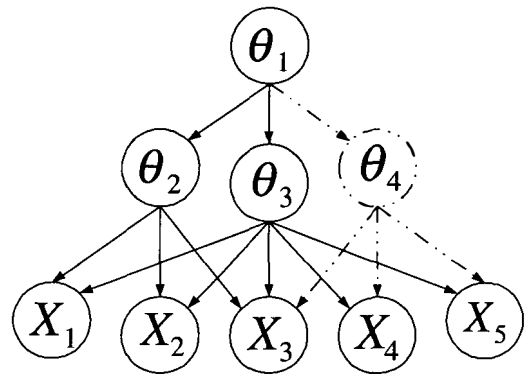

Figure 2: Node Exclusion Error Model

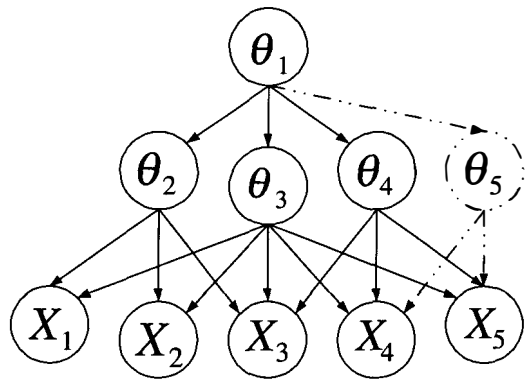

Figure 3: Node Inclusion Error Model

---

[7] By transposing the matrix of values it would be possible to utilize this procedure to evaluate the person-model fit rather than the model-data fit.

[8] The change from the Data Generation model is illustrated in this and subsequent figures by dashed lines--in this case showing the former placement of the now-excluded Treatment Planning node and associated directed edges.

### 3.4.2 Directed Edge Errors

The second stage of the study investigated the erroneous exclusion and inclusion of edges. Each was evaluated in two degrees: strong edge and weak edge. The weak edge and strong edge exclusion error models are given in Figures 4 and 5 respectively. The edge inclusion error model is shown in Figure 6; the strength of the spuriously added edge was determined by assigning strong or weak conditional probabilities to it. Again, all errors were centered on the $\theta_4$ node in the latent structure.

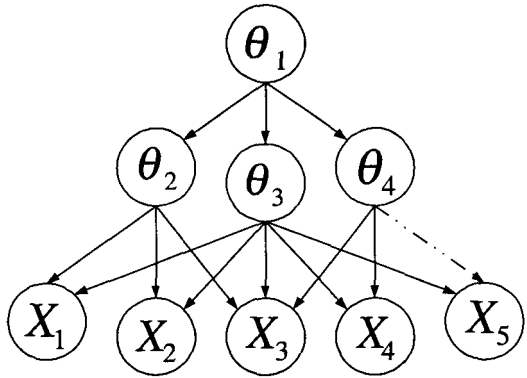

Figure 4: Weak Edge Exclusion Error Model

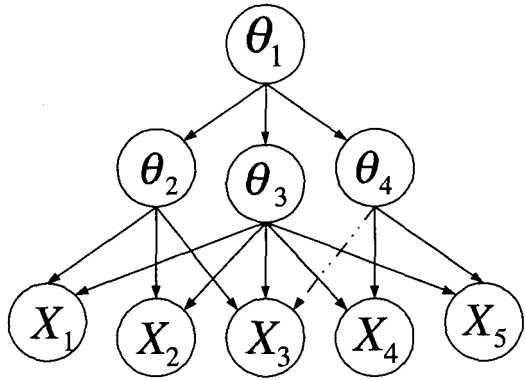

Figure 5: Strong Edge Exclusion Error Model

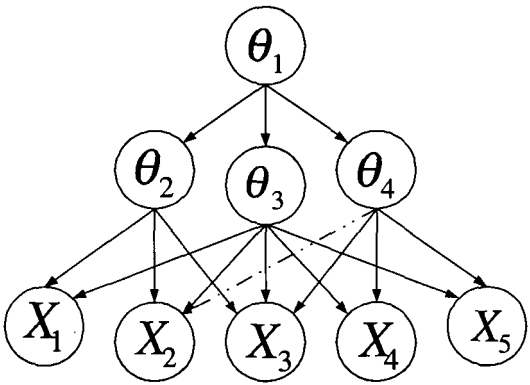

Figure 6: Edge (Strong or Weak) Inclusion Error Model

### 3.4.3 Variable State Errors

The third stage of the study investigated the erroneous exclusion or inclusion of a node state in a student model variable. The $\theta_4$ node was changed from its original three-state structure to a two-state structure (a state exclusion error), then to a four-state structure (a state inclusion error). These are relatively minor errors, since the two-state structure just collapsed two states into one, and the four-state structure was achieved by splitting one state into two with conditional probabilities interpolated from those of its neighboring states.

### 3.4.4 Prior Probability Error

The fourth stage of the study investigated the erroneous specification of prior probabilities in a latent variable[9]. The $\theta_4$ node was again the locus of the error, which altered the probabilities moderately from the ones in the data generation model.

## 3.5 PROCEDURE

Each stage of the study followed the same sequence of steps: 1) Generate a dataset (N=1000) consistent with the *posited model* (the model, either erroneous or true, that is the subject of model criticism). 2) Use the posited model to produce the probabilities (via 3.3) for each observable node for both the model-consistent data (from step 1) and the 'observed' data (from section 3.2). 3) Compute the fit indices (section 3.1) at various sample sizes for both the model-consistent data and the 'observed' data and determine the distributional properties of the indices. 4) Bootstrap[10] (Efron & Tibshirani, 1993) the model-consistent data (for posited model) to generate empirical distributions of values under the null hypothesis and determine critical values for evaluating the 'observed' data. 5) Evaluate the 'observed' data in light of these empirical distributions and critical values. This approach combines the methodology of the bootstrap with Rubin's (1984) use of frequency distributions to evaluate Bayesian models.

For each BN model (true and error models) this evaluation was conducted at sample sizes of 50, 100, 250, 500 and 1000 simulees. The larger sample sizes included the data from the smaller sample sizes. Each bootstrap data set had a sample size equal to that of the 'observed' data being evaluated, and critical values were established at the empirical values representing the 2.5% and 97.5% percentiles. This corresponds to a $p < .05$, two-tailed test.

---

[9] This technique could be applied in sensitivity analysis as well.

[10] The bootstrap is a means of approximating the distribution of a statistic by empirically calculating the statistic for numerous samples (with replacement) from the original sample. For this study 1000 samples of n=1000 were used to estimate the distribution (and critical values) for each statistic.

Values of the 'observed' data that exceeded these critical values were considered significant. A two-tailed test made it was possible to obtain significant results for better than expected model-data fit as well as misfit, though the latter is the primary concern of model criticism.

# 5 RESULTS

Plots of the resultant values for the global and node measures served as the first basis for evaluating the effectiveness of the model criticism indices. For each plot (examples are provided below) the x-axis indicates sample size (e.g. 50 indicates that the observed data and each of the 1,000 bootstrapped data sets had N=50) and the y-axis indicates the empirical value of the index. The dots connected by dashed lines represent the mean values for the 'observed' data and the solid lines represent the upper and lower critical values from the bootstrap (97.5% and 2.5% of the 1,000 bootstrapped data sets, respectively).

## 5.1 MODEL-DATA PLOTS

To illustrate typical trends in the results, we provide examples for the Ranked Probability Score as applied to the Data Generation and Node Exclusion models.

### 5.1.1 Global Measure Plots

Figure 7 shows the global measure results for the Data Generation model. Since the Data Generation model is known to be correct, significant deviations are false positive results. Small deviations from critical values occur at n's of 500 and 1000.

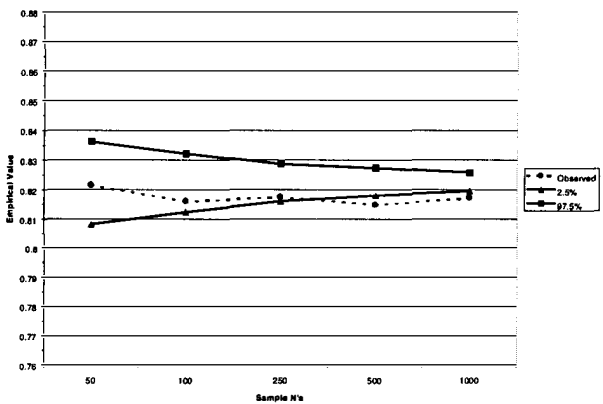

Figure 7: Global Measure Ranked Probability Score Results for the Data Generation Model

In contrast, the global measure results for the Node Exclusion model in Figure 8 demonstrate the deviations from a seriously misspecified model. Taken together, these results indicate the global measure has both specificity (Figure 7) and sensitivity (Figure 8) for identifying this particular latent structure error[11].

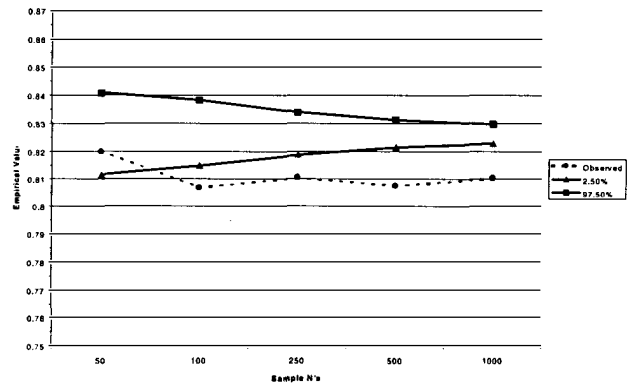

Figure 8: Global Measure Ranked Probability Score Results for the Node Exclusion Model

### 5.1.2 Node Measure Plots

The same approach was applied to the node measures. The node measure results for the Data Generation Model were predominantly within the bootstrap critical values, with an occasional value slightly beyond the cutoff. An example of such an occurrence is provided for the Patient 2 node as Figure 9.

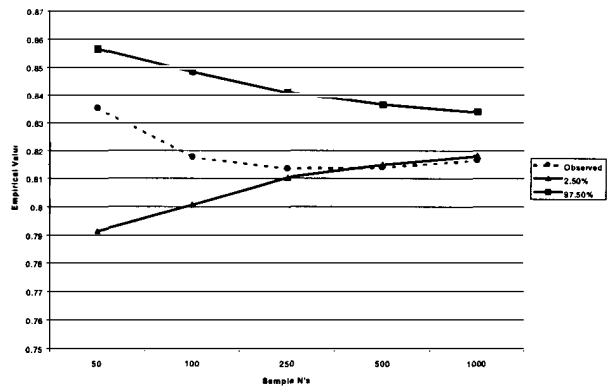

Figure 9: Patient 2 Node Measure Ranked Probability Score Results for the Data Generation Model

In contrast, nodes for observables closely associated with an error in the latent structure showed more dramatic deviations from the bootstrap distributions. Figure 10 provides an example, specifically the Patient 5 node measure for the Node Exclusion model.

---

[11] We also examined, in the same way, the Quadratic Brier Score (Brier, 1950), Good's Logarithmic Surprise Index (Good, 1954), and the Logarithmic Score (Cowell, Dawid, & Spiegelhalter, 1993). Higher rates of false positive significant results undermined their utility, so they are not discussed here.

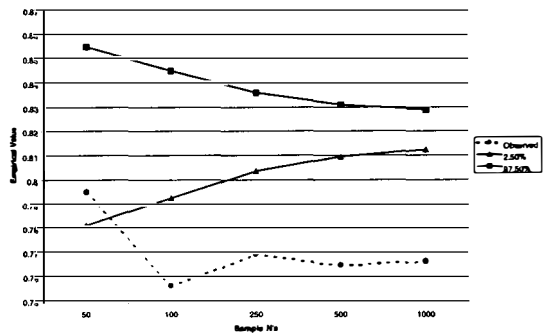

Figure 10: Patient 5 Node Measure Ranked Probability Score Results for the Node Exclusion Model

Three observations can be made about the application and utility of the node measures. First, the $x$ with the closest proximity and greatest degree of relationship to the source of the latent structure error was nearly always the first (by sample size) to identify model error, and produced the greatest degree of discrepancy from the bootstrap parameters. Second, nodes in close proximity but with weaker associations with the source of the error seldom deviated from the bootstrap distributions. Third, some nodes more distant from the location of the latent structure error produced significant deviations ("collateral significance"). However, the degree of deviation for such instances was always secondary and considerably less than for the first node identified.

Figure 11 shows an example of collateral significance, corresponding to the primary node of Figure 10, occurring for the node measure of Patient 1 in the Node Exclusion Model. Instances of collateral significance appeared for $x$s in models correct in their neighborhood but erroneous in other areas. In these cases, using $x$s which were modeled most incorrectly produced distorted predictive distributions for $x$s which were modeled correctly. Node indices can thus indicate the presence of a problem with respect to a given observable, but ways of rectifying the problem are not limited to ones that just focus on that observable.

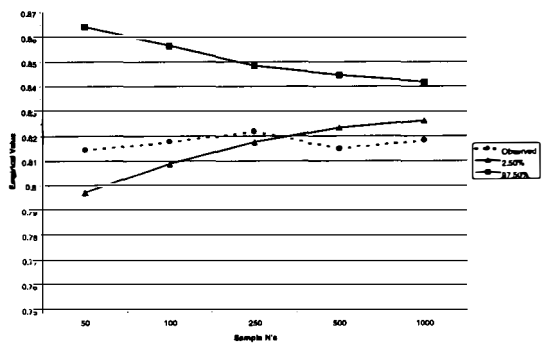

Figure 11: Patient 1 Node Measure Ranked Probability Score Results for the Node Exclusion Model

## 5.2 PLOT SUMMARIES

Table 2 summarizes the plots produced for the Ranked Probability Score. The 'Model' column indicates true or error model for which results are presented. The column marked 'Global' indicates the global measure results, and the columns marked 'Patient 1' through 'Patient 5' are the results for the $x$s. Numeric values in a cell indicate that at least one analysis (of the five sample sizes utilized) produced a significant deviation from the bootstrap distributions. The numeric values indicate which sample sizes produced significant deviations. Bold type represents cells where there was an error in the latent structure of the immediate parent variable, and a bold X appears in cells where there was an undetected error in the latent structure of the immediate parent. Cross-referencing the data in Table 2 to Figures 7 through 11 helps to clarify its interpretation. Tables 3 and 4 give similar summaries for Weaver's Surprise Index and Good's Logarithmic Score.

Table 2: Plot Summary for the Ranked Probability Score

| Model | Level/Node | | | | | |
|---|---|---|---|---|---|---|
| | Global | P1 | P2 | P3 | P4 | P5 |
| Data Generation | | | | | | |
| Node Exclusion | 100, 250, 500, 1000 | 500, 1000 | | **X** | **X** | 100, 250, 500, 1000 |
| Node Inclusion | | | | | **X** | 100, 250, 500, 1000 |
| State Exclusion | | | | **X** | **1000** | 500, 1000 |
| State Inclusion | | | | **X** | **X** | **X** |
| Prior Probability | | | | **X** | **X** | 500, 1000 |
| Strong Edge Exclusion | 100, 250, 500, 1000 | 500, 1000 | | | | 100, 250, 500, 1000 |
| Strong Edge Inclusion | 500, 1000 | | **250, 500, 1000** | | | 500, 1000 |
| Weak Edge Exclusion | | | | **X** | | 100, 250, 500, 1000 |
| Weak Edge Inclusion | | | **X** | | | |

Table 3: Plot Summary for Weaver's Surprise Index

| Model | Level/Node | | | | | |
|---|---|---|---|---|---|---|
| | Global | P1 | P2 | P3 | P4 | P5 |
| Data Generation | | | | | | |
| Node Exclusion | 100, 250, 500, 1000 | | | X | X | 100, 250, 500, 1000 |
| Node Inclusion | | | | | X | 500, 1000 |
| State Exclusion | | | X | 1000 | X | |
| State Inclusion | | | X | X | X | |
| Prior Probability | | | X | X | X | |
| Strong Edge Exclusion | 250, 500, 1000 | 1000 | | | | 100, 250, 500, 1000 |
| Strong Edge Inclusion | 500, 1000 | | 500, 1000 | | | 100, 250, 500, 1000 |
| Weak Edge Exclusion | | | 500, 1000 | | | |
| Weak Edge Inclusion | | X | | | | |

Table 4: Plot Summary for Good's Logarithmic Score

| Model | Level/Node | | | | | |
|---|---|---|---|---|---|---|
| | Global | P1 | P2 | P3 | P4 | P5 |
| Data Generation | | | | | | |
| Node Exclusion | | | | X | X | X |
| Node Inclusion | | | | | X | X |
| State Exclusion | | | | X | 100, 250, 500, 1000 | X |
| State Inclusion | | | | X | 100, 250, 500, 1000 | X |
| Prior Probability | | | | X | X | X |
| Strong Edge Exclusion | | | | | | X |
| Strong Edge Inclusion | | 100 | X | | | |
| Weak Edge Exclusion | | | | X | | 1000 |
| Weak Edge Inclusion | | | X | | | |

## 5 DISCUSSION

### 5.1 IMPLICATIONS

These results offer promise of utility for the Ranked Probability Score and Weaver's Surprise Index as global measures and node measures to detect specific types of modeling errors in the latent structure of BNs. For global measures, major error types (node exclusions and strong edge errors) in the latent structure were detectable. For node measures (preferably used in combination) these indices helped identify major latent structure errors (node errors and strong edge errors) at moderate sample sizes, and minor latent structure errors (weak edge errors, node state errors, and prior probability errors) at large sample sizes. The results suggest utility as node measures even in the absence of model-data misfit for global measures.

Furthermore, these results suggest that as node measures these indices can identify nodes in close proximity to the latent structure error, providing the modeler some direction for appropriate modification to the student model. This capability is complicated by the possibility of collateral significance of node measures. However, examining correlations among nodes, obtained for example by exercising the network, would allow the modeler to exploit collateral significance by knowing which latent nodes have strong associations with the observable node in question (one cannot tell whether a significant finding is direct or collateral!).

These results also suggest that Good's Logarithmic Score can be used as a node measure to detect errors of node state inclusion or exclusion in the latent structure of BNs. This finding may be of particular interest since neither of the other indices was able to detect these errors.

To the extent that these results generalize to other such BN models with latent variables, Table 5 suggests guidelines for the use of the Ranked Probability Score (RPS), Weaver's Surprise Index (WSI), and Good's Logarithmic Score (GLS) as node measures.

Table 5: Utilization as Node Measures

| N | Significant Deviation | | | Error Types |
|---|---|---|---|---|
| | GLS | RPS | WSI | |
| ≤250 | yes | no | no | node state exclusion; node state inclusion |
| | no | yes | no | node inclusion; strong edge inclusion |
| | no | yes | yes | node exclusion; strong edge exclusion |
| >250 and ≤1000 | yes | no | no | node state exclusion; node state inclusion |
| | no | yes | no | node state exclusion; prior probability error |
| | no | no | yes | weak edge exclusion |
| | no | yes | yes | node exclusion; node inclusion; strong edge exclusion; strong edge inclusion |

A key feature of the approach in this paper is the initial theory-driven investigation of the domain of interest (e.g. CTA) to inform the construction of a theoretical model, which is then subjected to empirical model criticism. This contrasts with the approach of Heckerman and colleagues (e.g. Heckerman, Geiger, & Chickering, 1995; Geiger, Heckerman & Meek, 1996), which utilizes Markov-chain Monte Carlo techniques to identify an optimally-fitting model purely empirically without a driving theoretical rationale, and then attempts to discern causal relationships within the resulting structure.

## 5.2 CAUTIONS AND LIMITATIONS

There are, of course, some cautions and limitations that must be recognized before widespread application of this methodology. The most important limitation is that we have demonstrated its utility in the context of a particular family of BN models. The results may vary for other BNs with different structure, size, associations among latent and observable variables, error types, error locations, and prior probabilities from the ones implemented in this study.

Also, this study utilized a two-tailed approach to implementing the bootstrap techniques while the distribution of index values (approximating a chi-square) and the interest in detecting model misfit rather than model overfit suggest that a one-tailed approach would be more appropriate.

Another limitation is that the node measures were evaluated without correcting for multiple tests. To control the Type I error rate at .05 a per-family-error-rate correction should be implemented to maximize power while maintaining the Type I error rate (a Bonferroni adjustment would be too conservative for such correlated observable nodes).

The preceding two limitations mitigate each other as one errs in a conservative direction while the other errs in a liberal fashion. A reanalysis was conducted on two of the models correcting these limitations to see if there would be any effect on the results and interpretation. The results strongly suggest that there would be no significant impact on the results or interpretation of this study. However, there is one notable sign of an advantage in these corrections: the near elimination of deviations from bootstrap parameters under the Data Generation model (null hypothesis) and corresponding node measures without ancestral errors in latent structure. This may have implications for future applicability of indices[12] that showed high false positive values.

---

[12] The Quadratic Brier Score (Brier, 1950), Good's Logarithmic Surprise Index (Good, 1954), and the Logarithmic Score (Cowell, Dawid, & Spiegelhalter, 1993)

## 5.3 FUTURE DIRECTIONS

Obviously an important direction for further research is to establish the generalizability of these results to BNs with latent variables by systematically manipulating BN features such as network size, associations, proportion of latent to observable nodes, etc. to determine whether model criticism is affected by such variations. In such investigations it may become apparent that there are variations in the efficacy of each of these indices studied in detecting various types of errors under various types of BN conditions.

## 6 CONCLUSION

The introduction of this methodology, and more critically, the emphasis on model criticism of BNs with latent variables in general, provides a means of maximizing the accuracy and utility of BN models for a variety of applications. As methods of providing empirical support or criticism of student models in cognitive assessment, these results provide a means of ensuring that the student models developed are appropriate representations of the constellation of knowledges, processes, and strategies which contribute to task performance. This capability offers the potential of helping the analyst to create a student model from a CTA by comparing modeled structures with preliminary performance data; to revise BN structures to improve classification decisions for examinees; to provide validity evidence for the student model in the substantive domain; and to identify examinees who do not fit the model. With such applications these indices would contribute to the production of more accurate cognitive models in less time, facilitate the implementation of BN and related methodologies in future applications, and support the construct validity of the resultant assessments and intelligent tutoring systems.

### Acknowledgements

The authors would like to thank Charlie Lewis for his suggestions regarding appropriate corrections for repeated node measures.